\documentclass[11pt,a4paper]{article}
\usepackage[hyperref]{acl2019}
\usepackage{times,subcaption}
\usepackage{latexsym}
\usepackage{amsmath,amsthm,amssymb,pstricks,pst-node}
\usepackage{url,xcolor}
\usepackage{graphicx}
\usepackage{tikz}
\usepackage{booktabs}
\usepackage{xspace,textcomp}
\usetikzlibrary{calc,trees,positioning,arrows,chains,shapes.geometric,%
    decorations.pathreplacing,decorations.pathmorphing,shapes,%
    matrix,shapes.symbols}

\tikzset{suppress join/.code={\def\tikz@after@path{}}}

\tikzset{
>=stealth',
  punktchain/.style={
    rectangle, 
    rounded corners, 
    draw=black, thin,
    text width=8em, 
    minimum height=3em, 
    text centered,
    on chain},
  circlechain/.style={
    circle,  
    draw=black, thin,
    text centered, 
    minimum size=3em,
    inner sep=1pt,
    on chain},
  gen_circle/.style={
    circle,  
    draw=black, thin,
    text centered, 
    minimum size=3em,
    inner sep=1pt,
    on chain},
  line/.style={draw, thick, <-},
  side/.style={
    rectangle,
    minimum width=1em,
    draw=black, thin,
    text width=2em, 
    minimum height=2em, 
    text centered},
  sidecircle/.style={
    circle,  
    fill=green!10,
    draw=black, thin,
    text centered, 
    minimum size=0.1em,
    inner sep=1pt,
    on chain},
  every join/.style={->, thin,shorten >=1pt},
  decoration={brace},
  tuborg/.style={decorate},
  tubnode/.style={midway, right=2pt},
  bignode/.style={font={\fontsize{23}{25}\selectfont}},
  regularnode/.style={font={\fontsize{18}{25}\selectfont}},
  smallnode/.style={font={\fontsize{10}{12}\selectfont}},
}
\aclfinalcopy 
\def\aclpaperid{2355} 

\newcommand{\kld}{\mathit{KL}}
\newcommand{\vmf}{\text{vMF}}

\newcommand{\vgvae}{VGVAE\xspace}
\newcommand{\prl}{PRL\xspace}
\newcommand{\wpl}{WPL\xspace}
\newcommand{\wn}{WN\xspace}
\newcommand{\lc}{LC\xspace}

\newcommand{\sted}{ST\xspace}

\newcommand{\decinit}{\textsc{init}\xspace}
\newcommand{\decswap}{\textsc{swap}\xspace}
\newcommand{\decconcat}{\textsc{concat}\xspace}

\newsavebox\FrameBox
\newenvironment{Frame}{%
   \par\setbox\FrameBox\hbox\bgroup\minipage{0.45\textwidth}\parskip\baselineskip\ignorespaces
}{%
   \endminipage\egroup\fbox{\box\FrameBox}\par
}

\title{Controllable Paraphrase Generation with a Syntactic Exemplar}

\author{Mingda Chen\qquad Qingming Tang\qquad Sam Wiseman\qquad Kevin Gimpel\\
Toyota Technological Institute at Chicago, Chicago, IL, 60637, USA\\
  {\tt \{mchen,qmtang,swiseman,kgimpel\}@ttic.edu}\\}

\date{}

\begin{document}
\maketitle
\begin{abstract}
Prior work on controllable text generation usually assumes that the controlled attribute can take on one of a small set of values known a priori. In this work, we propose a novel task, where the syntax of a generated sentence is controlled rather by a sentential exemplar. To evaluate quantitatively with standard metrics, we create a novel dataset with human annotations. We also develop a variational model with a neural module specifically designed for capturing syntactic knowledge and several multitask training objectives to promote disentangled representation learning. Empirically, the proposed model is observed to achieve improvements over baselines and learn to capture desirable characteristics.\footnote{Code and data are available at~\href{https://github.com/mingdachen/syntactic-template-generation}{\nolinkurl{github.com/mingdachen/syntactic-template-generation}}}
\end{abstract}
\section{Introduction}

Controllable text generation has recently become an area of intense focus in the natural language processing (NLP) community. Recent work has focused both on generating text satisfying certain stylistic requirements such as being formal or exhibiting a particular sentiment~\citep{hu17control,shen2017style,ficler2017controlling}, as well as on generating text meeting structural requirements, such as conforming to a particular template~\citep{iyyer2018adversarial,wiseman2018templates}.

These systems can be used in various application areas, such as text summarization~\cite{fan2018control}, adversarial example generation~\cite{iyyer2018adversarial}, dialogue~\cite{niu2018polite}, and data-to-document generation~\citep{wiseman2018templates}. However, prior work on controlled generation has typically assumed a known, finite set of values that the controlled attribute can take on. In this work, we are interested instead in the novel setting where the generation is controlled through an exemplar sentence (where any syntactically valid sentence is a valid exemplar).
We will focus in particular on using a sentential exemplar to control the syntactic realization of a generated sentence. This task can benefit  natural language interfaces to information systems by suggesting alternative invocation phrases for particular types of queries~\citep{kumar2017just}.
It can also bear on dialogue systems that seek to generate utterances that fit particular functional categories~\cite{ke-etal-2018-generating,li2019insufficient}.

To address this task, we propose a deep generative model with two latent variables, which are designed to capture semantics and syntax.
To achieve better disentanglement between these two variables, we design multi-task learning objectives that make use of paraphrases and word order information.
To further facilitate the learning of syntax, we additionally propose to train the syntactic component of our model with word noising and latent word-cluster codes. Word noising randomly replaces word tokens in the syntactic inputs based on a part-of-speech tagger used only at training time. Latent codes create a bottleneck layer in the syntactic encoder, forcing it to learn a more compact notion of syntax. The latter approach also learns interpretable word clusters. Empirically, these learning criteria and neural architectures lead to better generation quality and generally better disentangled representations.

\begin{figure}
    \small
    \centering
    \begin{subfigure}[t]{16.55cm}
        \begin{Frame}
        $X$: {\color{red}his teammates' eyes got an ugly, hostile expression.}\\
        $Y$: {\color{blue}the smell of flowers was thick and sweet.}\\
        $Z$: the eyes of his teammates had turned ugly and hostile.
        \end{Frame}
    \end{subfigure}
    \begin{subfigure}[t]{16.55cm}
        \begin{Frame}
        $X$: {\color{red}we need to further strengthen the agency's capacities.}\\
        $Y$: {\color{blue}the damage in this area seems to be quite minimal.}\\
        $Z$: the capacity of this office needs to be reinforced even further.
        \end{Frame}
    \end{subfigure}
    \caption{Examples from our annotated evaluation dataset of paraphrase generation using semantic input $X$ (red), syntactic exemplar $Y$ (blue), and the reference output $Z$ (black).
    }
    \label{example}
\end{figure}

To evaluate this task quantitatively, we manually create an evaluation dataset containing triples of a semantic exemplar sentence, a syntactic exemplar sentence, and a reference sentence incorporating the semantics of the semantic exemplar and the syntax of the syntactic exemplar. This dataset is created
by first automatically finding syntactic exemplars and then heavily editing them by ensuring (1) semantic variation between the syntactic inputs and the references, (2) syntactic similarity between the syntactic inputs and the references, and (3) syntactic variation between the semantic input and references.
Examples are shown in Figure~\ref{example}.
This dataset allows us to evaluate different approaches quantitatively using standard metrics, including BLEU~\cite{papi2002bleu} and ROUGE~\cite{lin2004rouge}. As the success of controllability of generated sentences also largely depends on the syntactic similarity between the syntactic exemplar and the reference, we propose a ``syntactic similarity'' metric based on evaluating tree edit distance between constituency parse trees of these two sentences after removing word tokens.

Empirically, we benchmark the syntactically-controlled paraphrase network (SCPN) of \citet{iyyer2018adversarial} on this novel dataset, which shows strong performance with the help of a supervised parser at test-time but also can be sensitive to the quality of the parse predictor. We show that using our word position loss effectively characterizes syntactic knowledge, bringing consistent and sizeable improvements over syntactic-related evaluation. The latent code module learns interpretable latent representations. Additionally, all of our models can achieve improvements over baselines. Qualitatively, we show that our models do suffer from the lack of an abstract syntactic representation, though we also show that SCPN and our models  exhibit similar artifacts.

\section{Related Work}
We focus primarily on the task of paraphrase generation, which
has received significant recent attention~\citep{quirk2004mono,prakash2016neural,mallinson2017paraphrase,dong2017learn,ma2018query,li2018paraphrase}. In order to disentangle the syntactic and semantic aspects of paraphrase generation we learn an explicit latent variable model using a variational autoencoder (VAE)~\citep{Kingma2014}, which is now commonly applied to text generation~\citep{bowman2016gen,miao2016neural,seme2017hybrid,serban2017piecewise,xu2018spherical,shen2019generating}.

In seeking to control generation with exemplars, our approach relates to recent work in controllable text generation. Whereas much work on controllable text generation seeks to control distinct attributes of generated text (e.g., its sentiment or formality)~\citep[\emph{inter alia}]{hu17control,shen2017style,ficler2017controlling,fu2018style,zhao2018adversarially,fan2018control}, there is also recent work which attempts to control structural aspects of the generation, such as its latent~\citep{wiseman2018templates} or syntactic~\citep{iyyer2018adversarial} template.

Our work is closely related to this latter category, and to the syntactically-controlled paraphrase generation of~\citet{iyyer2018adversarial} in particular, but our proposed model is different in that it simply uses a single \textit{sentence} as a syntactic exemplar rather than requiring a supervised parser. This makes our setting closer to style transfer in computer vision, in which an image is generated that combines the content from one image and the style from another~\citep{gatys2016image}. In particular, in our setting, we seek to generate a sentence that combines the semantics from one sentence with the syntax from another, and so we only require a pair of (unparsed) sentences. We also note recent, concurrent work that attempts to use sentences as exemplars in controlling generation~\citep{wang2019toward} in the context of data-to-document generation~\citep{wiseman2017challenges}.

Another related line of work builds generation upon sentential exemplars ~\cite{guu-etal-2018-generating,weston-etal-2018-retrieve,pandey-etal-2018-exemplar,cao-etal-2018-retrieve,haopeng-text-19} in order to improve the quality of the generation itself, rather than to allow for control over syntactic structures.

There has been a great deal of work in applying multi-task learning to improve performance on NLP tasks~\cite[\emph{inter alia}]{plank2016multilingual,rei2017semi,augen2017multitask,bollmann18multitask}. Some recent work used multi-task learning as a way of improving the quality or disentanglement of learned representations~\cite{zhao2017learning,goyal2017z,du2018variational,john2018disentangled}.

Part of our evaluation involves assessing the different characteristics captured in the semantic and syntactic encoders, relating them to work on learning disentangled representations in NLP, including morphological reinflection~\cite{zhou2017multi}, sequence labeling~\cite{chen2018vsl}, and sentence representations~\cite{mchen-multitask-19}.

\section{Methods}

\begin{figure}
    \centering
    \begin{tikzpicture}
  [node distance=1.5em,
  start chain=going right,
  scale=0.5,
  every node/.style={scale=0.5}]
	 \node[bignode, gen_circle] (x1) {$x$};
	 \node[right=2.0em of x1] (in) {};
	 \node[bignode, gen_circle, above=0.4em of in] (z) {$z$};
	 \node[bignode, gen_circle, below=0.4em of in] (y) {$y$};
	 \node[bignode, gen_circle, right=2.0em of in] (x2) {$x$};
	 
	 \draw[->,thin] (y.10) -> (x2.220);
	 \draw[->,thin] (z.350) -> (x2.140);

	 \draw[->,thin, dashed] (x1.50) -> (z.180);
	 \draw[->,thin, dashed] (x1.320) -> (y.160);
\end{tikzpicture}
    \caption{Graphical model. Dashed lines indicate the inference model. Solid lines indicate the generative model.}
    \label{fig:graph}
    \vspace{-2.0mm}
\end{figure}
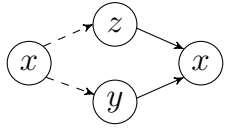

Given two sentences $X$ and $Y$,
our goal is to generate a sentence $Z$ that follows the syntax of $Y$ and the semantics of $X$.
We refer to $X$ and $Y$ as the semantic template and syntactic template, respectively.

To solve this problem, we follow \citet{mchen-multitask-19} and take an approach based on latent-variable probabilistic modeling, neural variational inference, and multi-task learning.
In particular, we assume a generative model that has two latent variables: $y$ for semantics and $z$ for syntax (as depicted in Figure~\ref{fig:graph}). We refer to our model as a vMF-Gaussian Variational Autoencoder (\vgvae). Formally, following the conditional independence assumptions in the graphical model,
the joint probability $p_\theta(x,y,z)$ can be factorized as:
\begin{align*}
    p_{\theta}(x, y, z) &= p_{\theta}(y) p_{\theta}(z) p_{\theta}(x \, | \, y, z) \\
    &= p_{\theta}(y) p_{\theta}(z) \prod_{t=1}^T p_\theta(x_t \, | \, x_{1:t-1}, y, z),
\end{align*}
\noindent where $x_t$ is the $t$th word of $x$ and $p_\theta(x_t \, | \, x_{1:t-1}, y, z)$ is given by a softmax over a vocabulary of size $V$. Further details on the parameterization are given below.

When applying neural variational inference, we assume a factorized approximated posterior $q_\phi(y\vert x)q_\phi(z\vert x)=q_\phi(y,z\vert x)$, which has also been used in some prior work~\cite{zhou2017multi,chen2018vsl}. Learning in \vgvae maximizes a lower bound of marginal log-likelihood:
\begin{equation}
\begin{aligned}
    &\log p_\theta(x)\geq\mathop\mathbb{E}_{\substack{y\sim q_\phi(y\vert x)\\z\sim q_\phi(z\vert x)}}[\log p_\theta(x\vert~z,y)\\
    &-\log\frac{q_\phi(z\vert x)}{p_\theta(z)}
    -\log\frac{q_\phi(y\vert x)}{p_\theta(y)}]\\
    &=\mathop\mathbb{E}_{\substack{y\sim q_\phi(y\vert x)\\z\sim q_\phi(z\vert x)}}[\log p_\theta(x\vert z,y)]-\kld(q_\phi(z\vert x)\Vert p_\theta(z))\\
    &-\kld(q_\phi(y\vert x)\Vert p_\theta(y))
\end{aligned}
\label{eq:elbo}
\end{equation}

\subsection{Parameterization}

\paragraph{vMF Distribution.}
We choose a von Mises-Fisher (vMF) distribution for the $y$ (semantic) latent variable.
vMF can be regarded as a Gaussian distribution on a hypersphere with two parameters: $\mu$ and $\kappa$. $\mu\in\mathbb{R}^m$ is a normalized vector (i.e., $\Vert\mu\Vert_2=1$) defining the mean direction. $\kappa\in\mathbb{R}_{\geq 0}$ is often referred to as a concentration parameter analogous to the variance in a Gaussian distribution. We will assume $q_\phi(y\vert x)$ follows a vMF distribution and $p_\theta(y)$ follows the uniform distribution $\vmf(\cdot,0)$. We follow~\citet{davidson2018hyperspherical} and use an acceptance-rejection scheme to sample from the vMF distribution.

\paragraph{Gaussian Distribution.} We assume $q_\phi(z\vert x)$ follows a Gaussian distribution $\mathcal{N}(\mu_\beta(x),\text{diag}(\sigma_\beta(x)))$ and that the prior $p_\theta(z)$ is $\mathcal{N}(0,I_{d})$, where $I_{d}$ is a $d\times d$ identity matrix.

\paragraph{Encoders.} At test time, we want to have different combinations of semantic and syntactic inputs, which naturally suggests separate parameterizations for $q_\phi(y\vert x)$ and $q_\phi(z\vert x)$. Specifically, $q_\phi(y\vert x)$ is parameterized by a word averaging encoder followed by a three-layer feedforward neural network since it has been observed that word averaging encoders perform surprisingly well for semantic tasks~\citep{wieting-16-full}. $q_\phi(z\vert x)$ is parameterized by a bidirectional long short-term memory network~(LSTM; \citealp{hochreiter1997long}) also followed by a three-layer feedforward neural network, where we concatenate the forward and backward vectors produced by the biLSTM and then take the average of these vectors.

\paragraph{Decoders.}
\begin{figure}
    \centering
    \includegraphics[scale=0.6]{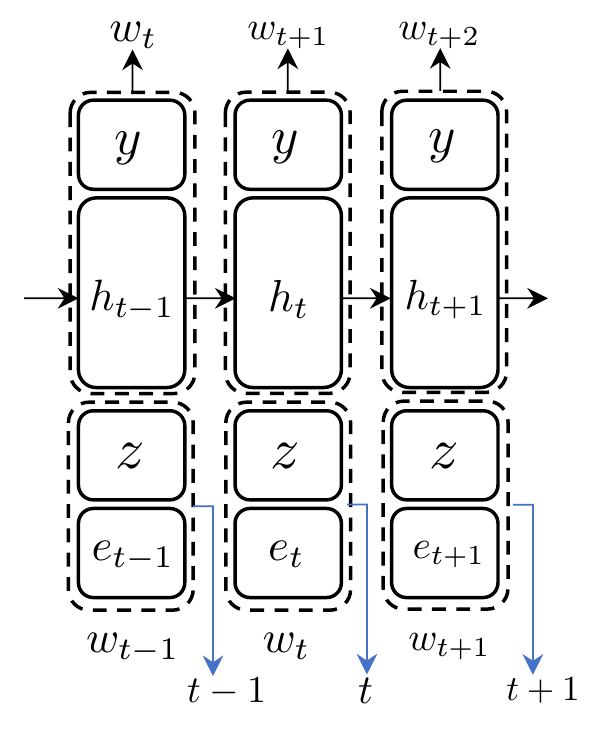}
    \caption{Diagram showing training of the decoder. Blue lines indicate the word position loss (\wpl).}
    \label{fig:decoder}
\end{figure}
As shown in Figure~\ref{fig:decoder}, at each time step, we concatenate the syntactic variable $z$ with the previous word's embedding as the input to the decoder and concatenate the semantic variable $y$ with the hidden vector output by the decoder for predicting the word at the next time step. Note that the initial hidden state of the decoder is always set to zero.

\subsection{Latent Codes for Syntactic Encoder}

Since what we want from the syntactic encoder is only the syntactic structure of a sentence, using standard word embeddings
tends to mislead the syntactic encoder to believe the syntax is manifested by the exact word tokens. An example is that the generated sentence often preserves the exact pronouns or function words in the syntactic input instead of making necessary changes based on the semantics. To alleviate this, we follow \citet{chen2018smaller} to represent each word with a latent code (\lc) for word clusters within the word embedding layer. Our goal is for this to create a bottleneck layer in the word embeddings, thereby forcing the syntactic encoder to learn a more abstract representation of the  syntax. However, since our purpose is not to reduce model size (unlike \citealp{chen2018smaller}), we marginalize out the latent code to get the embeddings during both training and testing. That is,
\begin{equation}
    e_w=\sum_{c_w}p(c_w)v_{c_w}\nonumber
\end{equation}
\noindent where $c_w$ is the latent code for word $w$, $v_{c_w}$ is the vector for latent code $c_w$, and $e_w$ is the resulting word embedding for word $w$. In our models, we use 10 binary codes produced by 10 feedforward neural networks based on a shared word embedding, and then we concatenate these 10 individual cluster vectors to get the final word embeddings.

\section{Multi-Task Learning}

We now describe several additional training losses designed to encourage a clearer separation of information in the semantic and syntactic variables. These losses were also considered in \cite{mchen-multitask-19}, but in the context of learning sentence representations.

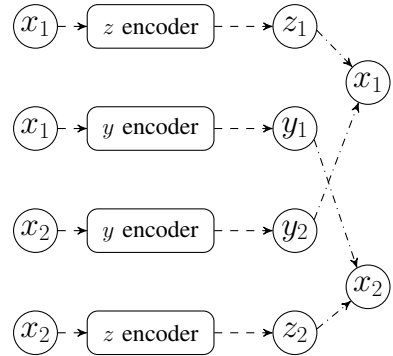
\begin{figure}
    \centering
    \begin{tikzpicture}
  [node distance=1.5em,
  start chain=going right,
  scale=0.5,
  every node/.style={scale=0.5}]
 \node[regularnode, punktchain] (e1) {$z$ encoder};
 \node[bignode, gen_circle, right=2.0em of e1] (z) {$z_1$};
 \node[bignode, gen_circle, left=1.0em of e1] (x1) {$x_1$};
 \node[regularnode, punktchain, below=2.0em of e1] (e2) {$y$ encoder};
 \node[bignode, gen_circle, right=2.0em of e2] (y) {$y_1$};
 \node[bignode, gen_circle, left=1.0em of e2] (x2) {$x_1$};

 \node[regularnode, punktchain, below=2.0em of e2] (e3) {$y$ encoder};
 \node[bignode, gen_circle, right=2.0em of e3] (z2) {$y_2$};
 \node[bignode, gen_circle, left=1.0em of e3] (x4) {$x_2$};
 \node[regularnode, punktchain, below=2.0em of e3] (e4) {$z$ encoder};
 \node[bignode, gen_circle, right=2.0em of e4] (y2) {$z_2$};
 \node[bignode, gen_circle, left=1.0em of e4] (x5) {$x_2$};

 \node[below=1.0em of e1] (middle) {};
 \node[bignode, gen_circle, right=6.5em of middle] (x3) {$x_1$};

 \draw[->,thin, dashed] (e1.0) -> (z.180);
 \draw[->,thin, dashed] (x1.0) -> (e1.180);

 \draw[->,thin, dashed] (e2.0) -> (y.180);
 \draw[->,thin, dashed] (x2.0) -> (e2.180);

 \%draw[->,thin] (z.350) -> (x3.135);

 \node[below=1.0em of e3] (middle) {};
 \node[bignode, gen_circle, right=6.5em of middle] (x6) {$x_2$};

 \draw[->,thin, dashed] (e3.0) -> (z2.180);
 \draw[->,thin, dashed] (x4.0) -> (e3.180);

 \draw[->,thin, dashed] (e4.0) -> (y2.180);
 \draw[->,thin, dashed] (x5.0) -> (e4.180);

 \draw[->,thin, dash dot] (y2.10) -> (x6.220);
 \draw[->,thin, dash dot] (z2.30) -> (x3.245);

 \draw[->,thin, dash dot] (y.330) -> (x6.120);
 \draw[->,thin, dash dot] (z.350) -> (x3.135);
\end{tikzpicture}
    \caption{Diagram showing the training process when using the paraphrase reconstruction loss (dash-dotted lines). The pair $(x_1, x_2)$ is a sentential paraphrase pair, the $y$'s are the semantic variables corresponding to each $x$, and the $z$'s are syntactic variables.}
    \label{fig:losses}
\end{figure}

\subsection{Paraphrase Reconstruction Loss}
Our first loss, the paraphrase reconstruction loss (PRL), requires a dataset of sentence paraphrase pairs.
The key assumption is that for a pair of paraphrastic sentences $x_1, x_2$, the semantics is shared but the syntax may differ. As shown in Figure~\ref{fig:losses}, we swap the paraphrases to the semantic encoder during training but keep the input to the syntactic encoder to be the same. It is defined as
\begin{equation}
\begin{aligned}
    \mathop\mathbb{E}_{\substack{y_2\sim q_\phi(y\vert x_2)\\z_1\sim q_\phi(z\vert x_1)}}[&\log p_\theta(x_1\vert y_2,z_1)] +\\ \mathop\mathbb{E}_{\substack{y_1\sim q_\phi(y\vert x_1)\\z_2\sim q_\phi(z\vert x_2)}}[&\log p_\theta(x_2\vert y_1,z_2)]
\end{aligned}
\label{eqn:prl}
\end{equation}
\noindent In the following experiments, unless explicitly noted, we will always include \prl as part of the model training and will discuss its effect in Section~\ref{sec:effect-prl}.

\subsection{Word Position Loss}
Since word ordering is relatively unimportant for semantic similarity~\cite{wieting-16-full}, we assume it is more relevant to the syntax of a sentence than to its semantics. Based on this, we introduce a word position loss (\wpl). As shown in Figure~\ref{fig:decoder}, \wpl is computed by predicting the position at each time step based on the concatenation of word embeddings with the syntactic variable $z$. That is,
\begin{equation}
    \text{WPL}\stackrel{\text{def}}{=\joinrel=}\mathop\mathbb{E}_{z\sim q_\phi(z\vert x)}\left[\sum_{t}\log\textrm{softmax}(f([e_t;z]))_t\right]
    \nonumber
\end{equation}
\noindent where $\textrm{softmax}(\cdot)_t$ indicates the probability at position $t$. Empirically, we observe that adding \wpl to both the syntactic encoder and decoder improves performance, so we always use it in our experiments unless otherwise indicated.

\section{Training}
\subsection{KL Weight}
As observed in previous work~\cite{alemi2016deep,bowman2016gen,higgins2016beta}, the weight of the KL divergence in Equation~\ref{eq:elbo} can be important when learning with latent variables. We attach weights to the KL divergence in Equation~\ref{eq:elbo} and tune them based on development set performance.

\subsection{Word Noising via Part-of-Speech Tags}
\begin{figure}
    \centering
    \includegraphics[scale=0.4]{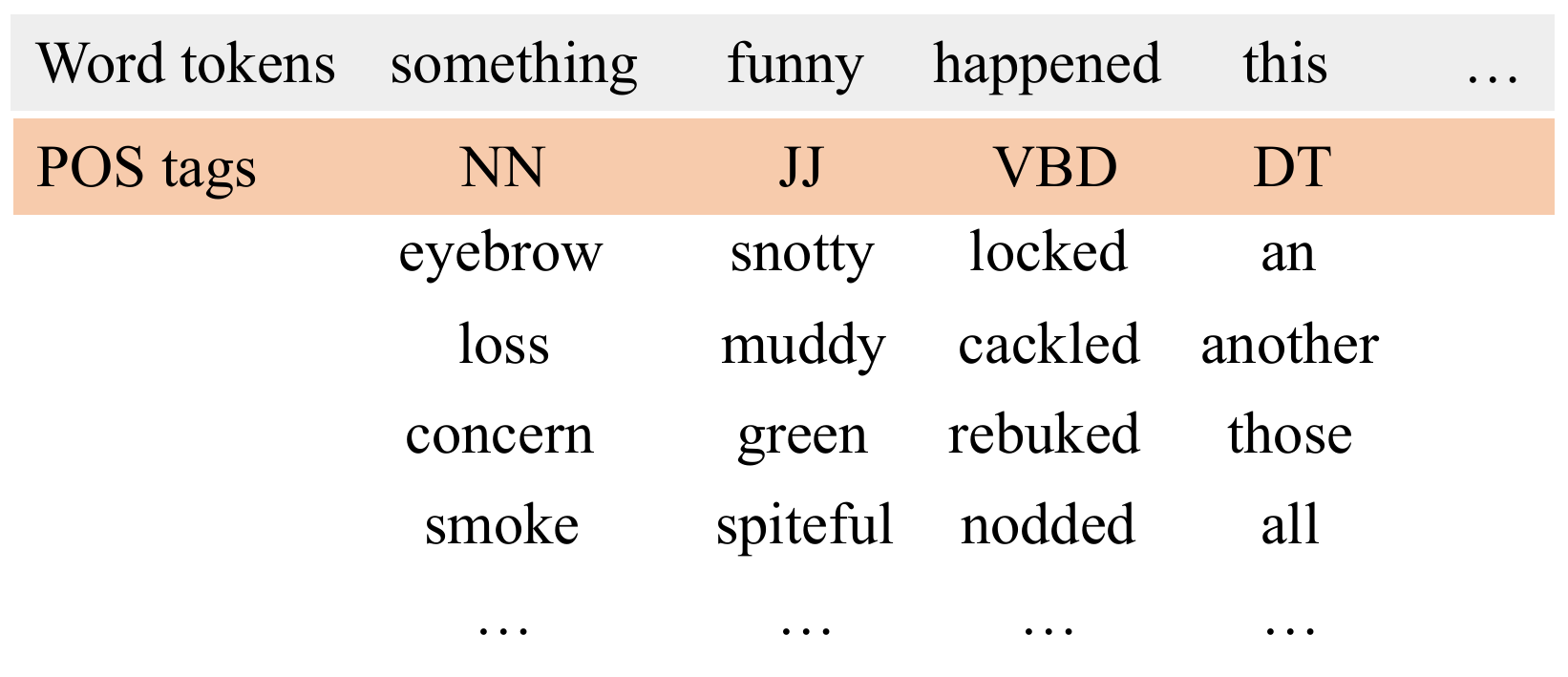}
    \caption{An example of word noising. For each word token in the training sentences, we randomly replace it with other words that share the same POS tags.
    }
    \label{fig:word-noise}
\end{figure}
In practice, we often observe that the syntactic encoder tends to remember word types instead of learning syntactic structures. To provide a more flexible notion of syntax, we add word noising (\wn) based on part-of-speech (POS) tags. More specifically, we tag the training set using the Stanford POS tagger~\cite{krist20013feature}. Then we group the word types based on the top two most frequent tags for each word type. During training, as shown in Figure~\ref{fig:word-noise}, we noise the syntactic inputs by randomly replacing word tokens based on the groups and tags we obtained. This provides our framework many examples of word interchangeability based on POS tags, and discourages the syntactic encoder from memorizing the word types in the syntactic input. When using \wn, the probability of noising a word is tuned based on development set performance.

\section{Experiments}

\subsection{Training Setup}
For training with the PRL, we require a training set of sentential paraphrase pairs. We use ParaNMT~\cite{para-nmt-acl-18}, a dataset of approximately 50 million paraphrase pairs.
To ensure there is enough variation between paraphrases, we filter out paraphrases with high BLEU score between the two sentences in each pair, which leaves us with around half a million paraphrases as our training set. All hyperparameter tuning is based on the BLEU score on the development set (see appendix for more details).

\begin{table*}[t]
\centering
\small
\begin{tabular}{l|cccccc}
\toprule
        & BLEU ($\uparrow$) & ROUGE-1 ($\uparrow$) & ROUGE-2 ($\uparrow$) & ROUGE-L ($\uparrow$) & METEOR ($\uparrow$) & ST ($\downarrow$) \\
\midrule
\multicolumn{7}{c}{Return-input baselines}\\
\midrule
Semantic input & 18.5 & 50.6 & 23.2 & 47.7 & 28.8 & 12.0 \\
Syntactic input & 3.3 & 24.4 & 7.5 & 29.1 & 12.1 & 5.9 \\
\midrule
\multicolumn{7}{c}{Our work}\\
\midrule
\vgvae              & 3.5  & 24.8 & 7.3  & 29.7 & 12.6 & 10.6 \\
\vgvae + \wpl       & 4.5  & 26.5 & 8.2  & 31.5 & 13.3 & 10.0 \\
\vgvae + \lc        & 3.3  & 24.0 & 7.2  & 29.4 & 12.5 & 9.1 \\
\vgvae + \lc + \wpl & 5.9  & 29.1 & 10.2 & 33.0 & 14.5 & 9.0 \\
\vgvae + \wn        & 13.0 & 43.2 & 20.2 & 47.0 & 23.8 & 6.8 \\
\vgvae + \wn + \wpl & 13.2 & 43.4 & 20.3 & 47.0 & 23.9 & 6.7 \\
\vgvae + \lc + \wn + \wpl & 13.6 & 44.7 & 21.0 & 48.3 & 24.8 & 6.7 \\
\midrule
\multicolumn{7}{c}{Prior work using supervised parsers}\\
\midrule
SCPN + template & 17.8 & 47.9 & 22.8 & 48.5 & 27.3 & 9.9 \\
SCPN + full parse & 19.2 & 50.4 & 26.1 & 53.5 & 28.4 & 5.9 \\
\bottomrule
\end{tabular}
\caption{Test results. The final metric (ST) measures the syntactic match between the output and the reference.
}
\label{test-set-res}
\end{table*}

\subsection{Evaluation Dataset and Metrics}
To evaluate models quantitatively, we manually annotate 1300 instances based on paraphrase pairs from ParaNMT independent from our training set. Each instance in the annotated data has three sentences: semantic input, syntactic input, and reference, where the semantic input and the reference can be seen as human generated paraphrases and the syntactic input shares its syntax with the reference but is very different from the semantic input in terms of semantics. The differences among these three sentences ensure the difficulty of this task. Figure~\ref{example} shows examples.

The annotation process involves two steps. We begin with a paraphrase pair $\langle u, v\rangle$. First, we use an automatic procedure to find, for each sentence $u$, a syntactically-similar but semantically-different other sentence $t$. We do this by seeking sentences $t$ with high edit distance of predicted POS tag sequences
and low BLEU score with $u$. Then we manually edit all three sentences to ensure (1) strong semantic match and large syntactic variation between the semantic input $u$ and reference $v$, (2) strong semantic match between the syntactic input $t$ and its post-edited version, and (3) strong syntactic match between the syntactic input $t$ and the reference $v$. We randomly pick 500 instances as our development set and use the remaining 800 instances as our test set. We perform additional manual filtering and editing of the test set to ensure quality.

For evaluation, we consider two categories of automatic evaluation metrics, designed to capture different components of the task. To measure roughly the amount of semantic content that matches between the predicted output and the reference, we report BLEU score (BL), METEOR score (MET; \citealp{ban2005meteor}) and three ROUGE scores, including ROUGE-1 (R-1), ROUGE-2 (R-2) and ROUGE-L (R-L). Even though these metrics are not purely based on semantic matching, we refer to them in this paper as ``semantic metrics'' to differentiate them from our second metric category, which we refer to as a ``syntactic metric''.
For the latter, to measure the syntactic similarity between generated sentences and the reference, we report the syntactic tree edit distance (\sted). To compute \sted, we first parse the sentences using Stanford CoreNLP~\cite{manning-EtAl:2014:P14-5}, and then compute the tree edit distance~\cite{zhang1989simple} between constituency parse trees after removing word tokens.

\subsection{Baselines}

We report results for three baselines. The first two baselines directly output the corresponding syntactic or semantic input for each instance. For the last baseline, we consider SCPN~\cite{iyyer2018adversarial}. As SCPN requires parse trees for both the syntactic and semantic inputs, we follow the process in their paper and use the Stanford shift-reduce constituency parser~\citep{manning-EtAl:2014:P14-5} to parse both,
then use the parsed sentences as inputs to SCPN. We report results for SCPN when using only the top two levels of the parse as input (template) and using the full parse as input (full parse).

\subsection{Results}

As shown in Table~\ref{test-set-res}, simply outputting the semantic input shows strong performance across the BLEU, ROUGE, and METEOR scores, which are more relevant to semantic similarity, but shows much worse performance in terms of \sted. On the other hand, simply returning the syntactic input leads to lower BLEU, ROUGE, and METEOR scores but also a very strong \sted score. These trends provide validation of the evaluation dataset, as they show that the reference and the semantic input match more strongly in terms of their semantics than in terms of their syntax, and also that the reference and the syntactic input match more strongly in terms of their syntax than in terms of their semantics. The goal in developing systems for this task is then to produce outputs with higher semantic metric scores than the syntactic input baseline and simultaneously higher syntactic scores than the semantic input baseline.

Among our models, adding \wpl leads to gains across both the semantic and syntactic metric scores. The gains are much larger without \wn, but even with \wn, adding \wpl improves nearly all scores.
Adding \lc typically helps the semantic metrics (at least when combined with WPL) without harming the syntactic metric (\sted).  We see the largest improvements, however, by adding \wn, which uses an automatic part-of-speech tagger at training time only. Both the semantic and syntactic metrics increase consistently with \wn, as the syntactic variable is shown many examples of word interchangeability based on POS tags.

While the SCPN yields very strong metric scores, there are several differences that make the SCPN results difficult to compare to those of our models. In particular, the SCPN uses a supervised parser both during training and at test time, while our strongest results merely require a POS tagger and only use it at training time.
Furthermore, since \sted is computed based on parse trees from a parser, systems that explicitly use constituency parsers at test time, such as SCPN, are likely to be favored by such a metric. This is likely the reason why SCPN can match the syntactic input baseline in  \sted.
Also, SCPN trains on a much larger portion of ParaNMT.

We find large differences in metric scores when SCPN only uses a parse template (i.e., the top two levels of the parse tree of the syntactic input). In this case, the results degrade, especially in \sted, showing that the performance of SCPN depends on the quality of the input parses.
Nonetheless, the SCPN results show the potential benefit of explicitly using a supervised constituency parser at both training and test time.
Future work can explore ways to combine syntactic parsers with our models for more informative training and more robust performance.

\section{Analysis}
\subsection{Effect of Multi-Task Training}
\paragraph{Effect of Paraphrase Reconstruction Loss.}\label{sec:effect-prl}
\begin{table}[t]
\setlength{\tabcolsep}{4pt}
\centering
\small
\begin{tabular}{l|c|c|c|c|c|c}
        & BL & R-1 & R-2 & R-L & MET & ST \\
\hline
\vgvae w/o \prl   & 2.0  & 23.4 & 4.3  & 26.4 & 11.3 & 11.8 \\
\vgvae w/ \prl    & 3.5  & 24.8 & 7.3  & 29.7 & 12.6 & 10.6
\end{tabular}
\caption{Test results when including \prl.}
\label{prl-res}
\vspace{-2.0mm}
\end{table}
We investigate the effect of \prl by removing \prl from training, which effectively makes \vgvae a variational autoencoder. As shown in Table~\ref{prl-res}, making use of pairing information can improve performance both in the semantic-related metrics and syntactic tree edit distance.

\paragraph{Effect of Position of Word Position Loss.}
\begin{table}[t]
\setlength{\tabcolsep}{4pt}
\centering
\small
\begin{tabular}{l|c|c|c|c|c|c}
        & BL & R-1 & R-2 & R-L & MET & ST \\
\hline
\vgvae w/o \wpl       & 3.5  & 24.8 & 7.3  & 29.7 & 12.6 & 10.6 \\
Dec. hidden state     & 3.6  & 24.9 & 7.3  & 29.7 & 12.6 & 10.5 \\
Enc. emb.             & 3.9  & 26.1 & 7.8  & 31.0 & 12.9 & 10.2 \\
Dec. emb.             & 4.1  & 26.3 & 8.1  & 31.3 & 13.1 & 10.1  \\
Enc. \&\ Dec. emb.      & 4.5  & 26.5 & 8.2  & 31.5 & 13.3 & 10.0
\end{tabular}
\caption{Test results with \wpl at different positions. }
\label{wpl-res}
\vspace{-2.mm}
\end{table}

We also study the effect of the position of \wpl by (1) using the decoder hidden state, (2) using the concatenation of word embeddings in the syntactic encoder and the syntactic variable, (3) using the concatenation of word embeddings in the decoder and the syntactic variable,  or (4) adding it on both the encoder embeddings and decoder word embeddings. Table~\ref{wpl-res} shows that adding \wpl on hidden states can help improve performance slightly but not as good as adding it on word embeddings. In practice, we also observe that the value of \wpl tends to vanish when using \wpl on hidden states, which is presumably caused by the fact that LSTMs have sequence information, making the optimization of \wpl trivial.
We also observe that adding \wpl to both the encoder and decoder brings the largest improvement.

\subsection{Encoder Analysis}

To investigate what has been learned in the encoder, we evaluate $q_\phi(y\vert x)$ and $q_\phi(z\vert x)$ on both semantic similarity tasks and syntactic similarity tasks and also inspect the latent codes.

\paragraph{Semantic Similarity.}

\begin{table}[t]
\setlength{\tabcolsep}{3pt}
\centering
\small
\begin{tabular}{l|c|c}
        & Semantic var. & Syntactic var. \\
\hline
\vgvae                    &   64.8 &   14.5 \\
\vgvae + \wpl             &   65.2 &   10.5 \\
\vgvae + \lc              &   67.2 &   29.0 \\
\vgvae + \lc + \wpl       &   67.9 &   8.5  \\
\vgvae + \wn              &   71.1 &   10.2  \\
\vgvae + \wn + \wpl       &   72.9 & 9.8 \\
\vgvae + \lc + \wn + \wpl &   74.3 & 7.4 \\
\end{tabular}
\caption{Pearson correlation (\%) for STS Benchmark test set.}
\label{encoder-sts}
\vspace{-2.0mm}
\end{table}

We use cosine similarity between two variables encoded by the inference networks as the predictions and then compute Pearson correlations on the STS Benchmark test set~\cite{cer2017semeval}. As shown in Table~\ref{encoder-sts}, the semantic variable $y$ always outperforms the  syntactic variable $z$ by a large margin, suggesting that different variables have captured different information. Every time when we add \wpl the differences in performance between the two variables increases.
Moreover, the differences between these two variables are correlated with the performance of models in Table~\ref{test-set-res}, showing that a better generation system has a more disentangled latent representation.

\paragraph{Syntactic Similarity.}

\begin{table}[t]
\setlength{\tabcolsep}{4pt}
\centering
\small
\begin{tabular}{l|c|c|c|c}
&\multicolumn{2}{c|}{Semantic var.} & \multicolumn{2}{c}{Syntactic var.} \\
   & $F_1$ & Acc. & $F_1$ & Acc.  \\
\hline
Random & 19.2 & 12.9 & - & - \\
Best   & 71.1 & 62.3 & - & - \\
\hline
\vgvae                &20.7 & 24.9 &25.9 & 28.8 \\
\vgvae + \wpl         &21.2 & 25.3 &31.1 & 33.3 \\
\vgvae + \lc          &21.6 & 25.5 &29.0 & 32.4  \\
\vgvae + \lc + \wpl   &18.9 & 23.5 &31.2 & 33.5 \\
\vgvae + \wn          &20.6 & 18.1 &28.4 & 30.4 \\
\vgvae + \wn + \wpl   &20.0 & 24.6 &43.7 & 40.8 \\
\vgvae + \lc +\wn + \wpl & 20.3 & 24.8 & 43.7 & 40.9\\
\end{tabular}
\caption{Labeled $F_1$ score (\%) and accuracy (\%) on syntactic similarity tasks from \citet{mchen-multitask-19}.}
\label{encoder-syntax}
\end{table}

We use the syntactic evaluation tasks from \citet{mchen-multitask-19} to evaluate the syntactic knowledge encoded in the encoder. The tasks are based on a 1-nearest-neighbor constituency parser or POS tagger. To understand the difficulty of these two tasks, Table~\ref{encoder-syntax} shows results for two baselines. ``Random'' means randomly pick candidates as predictions. The second baseline (``Best'') is to compute the pairwise scores between the test instances and the sentences in the candidate pool and then take the maximum values. It can be seen as the upper bound performance for these tasks.

As shown in Table~\ref{encoder-syntax}, similar trends are observed as in  Tables~\ref{test-set-res} and \ref{encoder-sts}. When adding \wpl or \wn, there is a boost in the syntactic similarity for the syntactic variable. Adding \lc also helps the performance of the syntactic variable slightly.

\paragraph{Latent Code Analysis.}

\begin{table}[t]
\setlength{\tabcolsep}{4pt}
\centering
\small
\begin{tabular}{l|c}
12 & does must could shall do wo 's did ai 'd 'll should \\\hline
451 & watching wearing carrying thrown refuse drew  \\ \hline
11 & ? : * $\gg$ ! ; ) . '' , '  \\ \hline
18 & maybe they because if where but we when how \\ \hline
41279 & elvish festive freeway anteroom jennifer terrors  \\ \hline
10 & well $\langle$unk$\rangle$ anyone okay now everybody someone \\ \hline
165 & supposedly basically essentially rarely officially  \\ \hline
59 & using on by into as the with within under quite \\ 
\end{tabular}
\caption{Examples of learned word clusters. Each row is a different clusters. Numbers in the first column indicate the number of words in that cluster.}
\label{lc-analysis}
\end{table}

We look into the learned word clusters by taking the argmax of latent codes and treating it as the cluster membership of each word. Although these are not the exact word clusters we would use during test time (because we marginalize over the latent codes), it provides us intuition on what individual cluster vectors have contributed to the final word embeddings. As shown in Table~\ref{lc-analysis}, the words in the first and last rows are mostly function words. The second row has verbs. The third row has special symbols. The fourth row also has function words but somewhat different from the first row. The fifth row is a large cluster populated by content words, mostly nouns and adjectives.
The sixth row has words that are not very important semantically and the seventh row has mostly adverbs. We also observe that the size of clusters often correlates with how strongly it relates to topics. In Table~\ref{lc-analysis}, clusters that have size under 20 are often function words while the largest cluster (5th row) has words with the most concrete meanings.

\begin{table}[t]
\setlength{\tabcolsep}{5pt}
\centering
\small
\begin{tabular}{l|c|c|c|c|c|c}
        & BL & R-1 & R-2 & R-L & MET & ST \\
\hline
\lc        & 13.6 & 44.7 & 21.0 & 48.3 & 24.8 & 6.7 \\
Single \lc & 12.9 & 44.2 & 20.3 & 47.4 & 24.1 & 6.9 \\
\end{tabular}
\caption{Test results when using a single code.}
\label{lc-varaints-res}
\end{table}

We also compare the performance of \lc by using a single latent code that has 50 classes. The results in Table~\ref{lc-varaints-res} show that it is better to use smaller number of classes for each cluster instead of using a cluster with a large number of classes.

\begin{table*}[t]
\setlength{\tabcolsep}{4pt}
\centering
\small
\begin{tabular}{|p{0.18\textwidth}|p{0.18\textwidth}|p{0.18\textwidth}|p{0.18\textwidth}|p{0.18\textwidth}|}
\hline
\multicolumn{1}{|c|}{Semantic input} & \multicolumn{1}{c|}{Syntactic input} & \multicolumn{1}{c|}{Reference} & \multicolumn{1}{c|}{SCPN + full parse} & \multicolumn{1}{c|}{Our best model} \\
\hline
\hline
don't you think that's a quite aggressive message? & that's worth something, ain't it? & that's a pretty aggressive message, don't you think? & that's such news, don't you? & that's impossible message, aren't you? \\\hline

if i was there, i would kick that bastard in the ass.& they would've delivered a verdict in your favor. & i would've kicked that bastard out on his ass. & you'd have kicked the bastard in my ass. & she would've kicked the bastard on my ass. \\\hline

with luck, it may turn out you're right. & of course, i could've done better. & if lucky, you will be proved correct. & with luck, i might have gotten better. & of course, i'll be getting lucky. \\\hline

they can't help, compassion is unbearable. & love is straightforward and it is lasting. & their help is impossible and compassion is insufferable. & compassion is unbearable but it is excruciating. & compassion is unacceptable and it is intolerable. \\\hline

her yelling sounds sad.  &  she looks beautiful. shining like a star. & she sounds sad. yelling like that. & she's sad. screaming in the air. & she sounds sad. screaming like a scream. \\\hline

me, scare him? & how dare you do such thing? & how can i scare him? & why do you have such fear? & why do you scare that scare?\\
\hline

\end{tabular}
\caption{Examples of generated sentences.
}
\label{gen-sents}
\end{table*}

\subsection{Effect of Decoder Structure}
\begin{figure}
    \centering
    \begin{subfigure}[t]{0.1\textwidth}
    \includegraphics[scale=0.6]{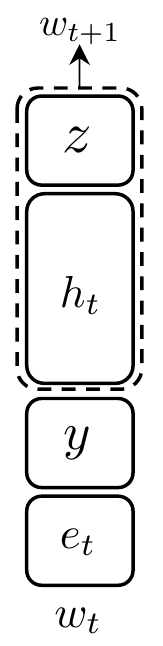}
    \end{subfigure}%
    \begin{subfigure}[t]{0.1\textwidth}
    \includegraphics[scale=0.6]{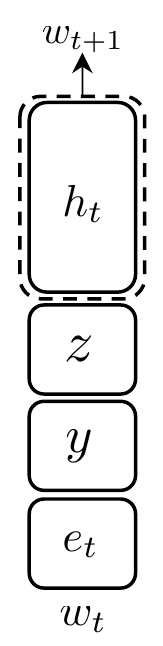}
    \end{subfigure}%
    \begin{subfigure}[t]{0.1\textwidth}
    \includegraphics[scale=0.6]{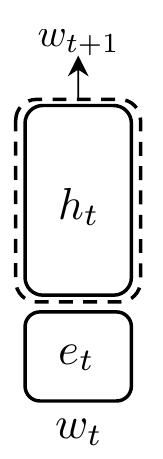}
    \end{subfigure}
    \caption{Variants of decoder. Left (\decswap): we swap the position of variable $y$ and $z$. Middle (\decconcat): we concatenate word embedding with $y$ and $z$ as input to decoder. Right (\decinit): we use word embeddings as input to the decoder and use the concatenation of $y$ and $z$ to compute the initial hidden state of the decoder.}
    \label{fig:decoder-variants}
\end{figure}

\begin{table}[t]
\setlength{\tabcolsep}{5pt}
\centering
\small
\begin{tabular}{l|c|c|c|c|c|c}
        & BL & R-1 & R-2 & R-L & MET & ST \\
\hline
\vgvae     & 4.5  & 26.5 & 8.2 & 31.5 & 13.3 & 10.0 \\
\decinit   & 3.5  & 22.7 & 6.0 & 24.9 & 9.8  & 11.5 \\
\decconcat & 4.0  & 23.9 & 6.6 & 27.9 & 11.2 & 10.9 \\
\decswap   & 4.3  & 25.6 & 7.5 & 30.4 & 12.5 & 10.5
\end{tabular}
\caption{Test results with decoder variants. }
\label{dec-variants-res}
\end{table}

As shown in Figure~\ref{fig:decoder-variants}, we evaluate three variants of the decoder, namely \decinit, \decconcat, and \decswap. For \decinit, we use the concatenation of semantic variable $y$ and syntactic variable $z$ for computing the initial hidden state of decoder and then use the word embedding as input and hidden state to predict the next word. For \decconcat, we move both $y$ and $z$ to the input of the decoder and use the concatenation of these two variables as input to the decoder and use the hidden state for predicting the next word. For \decswap, we swap the position of $y$ and $z$ to use the concatenation of $y$ and word embeddings as input to the decoder and the concatenation of $z$ and hidden states as output for predicting the next word. Results for these three settings are shown in Table~\ref{dec-variants-res}. \decinit performs the worst across the three settings. Both \decconcat and \decswap have variables in each time step in the decoder, which improves performance. \decswap arranges variables in different positions in the decoder and further improves over \decconcat in all metrics.

\subsection{Generated Sentences}

We show several generated sentences in Table~\ref{gen-sents}. We observe that both SCPN and our model suffer from the same problems. When comparing syntactic input and results from both our models and SCPN, we find that they are always the same length. This can often lead to problems like the first example in Table~\ref{gen-sents}. The length of the syntactic input is not sufficient for expressing the semantics in the semantic input, which causes the generated sentences from both models to end at ``you?'' and omit the verb ``think''. Another problem is in the consistency of pronouns between the generated sentences and the semantic inputs. An example is the second row in Table~\ref{gen-sents}. Both models alter ``i'' to be either ``you'' or ``she'' while the ``kick that bastard in the ass'' becomes ``kicked the bastard in my ass''.

We found that our models sometimes can generate nonsensical sentences, for example the last row in Table~\ref{gen-sents}.
while SCPN, which is trained on a much larger corpus, does not have this problem. Also, our models can sometimes be distracted by the word tokens in the syntactic input as shown in the 3rd row in Table~\ref{gen-sents}, where our model directly copies ``of course'' from the syntactic input while since SCPN uses a parse tree, it outputs ``with luck''. In some rare cases where the function words in both syntactic inputs and the references are the exactly the same, our models can perform better than SCPN, e.g.,
the last two rows in Table~\ref{gen-sents}. Generated sentences from our model make use of the word tokens ``and'' and ``like'' while SCPN does not have access to this information and generates inferior sentences.

\section{Conclusion}

We proposed a novel setting for controlled text generation, which does not require prior knowledge of all the values the control variable might take on. We also proposed a variational model accompanied with a neural component and multiple multi-task training objectives for addressing this task. The proposed approaches do not rely on a test-time parser or tagger and
outperform our baselines. Further analysis shows the model has learned both interpretable and disentangled representations.

\section*{Acknowledgments}
We would like to thank the anonymous reviewers, NVIDIA for donating GPUs used in this research,
and Google for a faculty research award to K.~Gimpel that partially supported this research.

\bibliography{acl2019}

\begin{thebibliography}{59}
\expandafter\ifx\csname natexlab\endcsname\relax\def\natexlab#1{#1}\fi

\bibitem[{Alemi et~al.(2017)Alemi, Fischer, Dillon, and Murphy}]{alemi2016deep}
Alexander~A. Alemi, Ian Fischer, Joshua~V. Dillon, and Kevin Murphy. 2017.
\newblock Deep variational information bottleneck.
\newblock In \emph{Proceedings of ICLR}.

\bibitem[{Augenstein and S{\o}gaard(2017)}]{augen2017multitask}
Isabelle Augenstein and Anders S{\o}gaard. 2017.
\newblock \href {https://doi.org/10.18653/v1/P17-2054} {Multi-task learning of
  keyphrase boundary classification}.
\newblock In \emph{Proceedings of the 55th Annual Meeting of the Association
  for Computational Linguistics (Volume 2: Short Papers)}, pages 341--346.
  Association for Computational Linguistics.

\bibitem[{Banerjee and Lavie(2005)}]{ban2005meteor}
Satanjeev Banerjee and Alon Lavie. 2005.
\newblock \href {http://aclweb.org/anthology/W05-0909} {{METEOR}: An automatic
  metric for {MT} evaluation with improved correlation with human judgments}.
\newblock In \emph{Proceedings of the ACL Workshop on Intrinsic and Extrinsic
  Evaluation Measures for Machine Translation and/or Summarization}, pages
  65--72. Association for Computational Linguistics.

\bibitem[{Bollmann et~al.(2018)Bollmann, S{\o}gaard, and
  Bingel}]{bollmann18multitask}
Marcel Bollmann, Anders S{\o}gaard, and Joachim Bingel. 2018.
\newblock \href {http://aclweb.org/anthology/W18-3403} {Multi-task learning for
  historical text normalization: Size matters}.
\newblock In \emph{Proceedings of the Workshop on Deep Learning Approaches for
  Low-Resource NLP}, pages 19--24. Association for Computational Linguistics.

\bibitem[{Bowman et~al.(2016)Bowman, Vilnis, Vinyals, Dai, Jozefowicz, and
  Bengio}]{bowman2016gen}
Samuel~R. Bowman, Luke Vilnis, Oriol Vinyals, Andrew Dai, Rafal Jozefowicz, and
  Samy Bengio. 2016.
\newblock \href {https://doi.org/10.18653/v1/K16-1002} {Generating sentences
  from a continuous space}.
\newblock In \emph{Proceedings of The 20th SIGNLL Conference on Computational
  Natural Language Learning}, pages 10--21. Association for Computational
  Linguistics.

\bibitem[{Cao et~al.(2018)Cao, Li, Li, and Wei}]{cao-etal-2018-retrieve}
Ziqiang Cao, Wenjie Li, Sujian Li, and Furu Wei. 2018.
\newblock \href {https://www.aclweb.org/anthology/P18-1015} {Retrieve, rerank
  and rewrite: Soft template based neural summarization}.
\newblock In \emph{Proceedings of the 56th Annual Meeting of the Association
  for Computational Linguistics (Volume 1: Long Papers)}, pages 152--161,
  Melbourne, Australia. Association for Computational Linguistics.

\bibitem[{Cer et~al.(2017)Cer, Diab, Agirre, Lopez-Gazpio, and
  Specia}]{cer2017semeval}
Daniel Cer, Mona Diab, Eneko Agirre, Inigo Lopez-Gazpio, and Lucia Specia.
  2017.
\newblock \href {https://doi.org/10.18653/v1/S17-2001} {Semeval-2017 task 1:
  Semantic textual similarity multilingual and crosslingual focused
  evaluation}.
\newblock In \emph{Proceedings of the 11th International Workshop on Semantic
  Evaluation (SemEval-2017)}, pages 1--14. Association for Computational
  Linguistics.

\bibitem[{Chen and Gimpel(2018)}]{chen2018smaller}
Mingda Chen and Kevin Gimpel. 2018.
\newblock \href {https://doi.org/10.18653/v1/N18-2116} {Smaller text
  classifiers with discriminative cluster embeddings}.
\newblock In \emph{Proceedings of the 2018 Conference of the North American
  Chapter of the Association for Computational Linguistics: Human Language
  Technologies, Volume 2 (Short Papers)}, pages 739--745. Association for
  Computational Linguistics.

\bibitem[{Chen et~al.(2018)Chen, Tang, Livescu, and Gimpel}]{chen2018vsl}
Mingda Chen, Qingming Tang, Karen Livescu, and Kevin Gimpel. 2018.
\newblock \href {http://aclweb.org/anthology/D18-1020} {Variational sequential
  labelers for semi-supervised learning}.
\newblock In \emph{Proceedings of the 2018 Conference on Empirical Methods in
  Natural Language Processing}, pages 215--226. Association for Computational
  Linguistics.

\bibitem[{Chen et~al.(2019)Chen, Tang, Wiseman, and
  Gimpel}]{mchen-multitask-19}
Mingda Chen, Qingming Tang, Sam Wiseman, and Kevin Gimpel. 2019.
\newblock \href {https://www.aclweb.org/anthology/N19-1254} {A multi-task
  approach for disentangling syntax and semantics in sentence representations}.
\newblock In \emph{Proceedings of the 2019 Conference of the North {A}merican
  Chapter of the Association for Computational Linguistics: Human Language
  Technologies, Volume 1 (Long and Short Papers)}, pages 2453--2464,
  Minneapolis, Minnesota. Association for Computational Linguistics.

\bibitem[{Davidson et~al.(2018)Davidson, Falorsi, De~Cao, Kipf, and
  Tomczak}]{davidson2018hyperspherical}
Tim~R. Davidson, Luca Falorsi, Nicola De~Cao, Thomas Kipf, and Jakub~M.
  Tomczak. 2018.
\newblock Hyperspherical variational auto-encoders.
\newblock \emph{34th Conference on Uncertainty in Artificial Intelligence
  (UAI-18)}.

\bibitem[{Dong et~al.(2017)Dong, Mallinson, Reddy, and Lapata}]{dong2017learn}
Li~Dong, Jonathan Mallinson, Siva Reddy, and Mirella Lapata. 2017.
\newblock \href {https://doi.org/10.18653/v1/D17-1091} {Learning to paraphrase
  for question answering}.
\newblock In \emph{Proceedings of the 2017 Conference on Empirical Methods in
  Natural Language Processing}, pages 875--886. Association for Computational
  Linguistics.

\bibitem[{Du et~al.(2018)Du, Li, He, Xu, Bing, and Wang}]{du2018variational}
Jiachen Du, Wenjie Li, Yulan He, Ruifeng Xu, Lidong Bing, and Xuan Wang. 2018.
\newblock Variational autoregressive decoder for neural response generation.
\newblock In \emph{Proceedings of the 2018 Conference on Empirical Methods in
  Natural Language Processing}, pages 3154--3163.

\bibitem[{Fan et~al.(2018)Fan, Grangier, and Auli}]{fan2018control}
Angela Fan, David Grangier, and Michael Auli. 2018.
\newblock \href {http://aclweb.org/anthology/W18-2706} {Controllable
  abstractive summarization}.
\newblock In \emph{Proceedings of the 2nd Workshop on Neural Machine
  Translation and Generation}, pages 45--54. Association for Computational
  Linguistics.

\bibitem[{Ficler and Goldberg(2017)}]{ficler2017controlling}
Jessica Ficler and Yoav Goldberg. 2017.
\newblock Controlling linguistic style aspects in neural language generation.
\newblock In \emph{Proceedings of the Workshop on Stylistic Variation}, pages
  94--104.

\bibitem[{Fu et~al.(2018)Fu, Tan, Peng, Zhao, and Yan}]{fu2018style}
Zhenxin Fu, Xiaoye Tan, Nanyun Peng, Dongyan Zhao, and Rui Yan. 2018.
\newblock Style transfer in text: Exploration and evaluation.
\newblock In \emph{{A}{A}{A}{I}}.

\bibitem[{Gatys et~al.(2016)Gatys, Ecker, and Bethge}]{gatys2016image}
Leon~A Gatys, Alexander~S Ecker, and Matthias Bethge. 2016.
\newblock Image style transfer using convolutional neural networks.
\newblock In \emph{Proceedings of the IEEE Conference on Computer Vision and
  Pattern Recognition}, pages 2414--2423.

\bibitem[{Goyal et~al.(2017)Goyal, Sordoni, C{\^o}t{\'e}, Ke, and
  Bengio}]{goyal2017z}
Anirudh Goyal Alias~Parth Goyal, Alessandro Sordoni, Marc-Alexandre
  C{\^o}t{\'e}, Nan~Rosemary Ke, and Yoshua Bengio. 2017.
\newblock Z-forcing: Training stochastic recurrent networks.
\newblock In \emph{Advances in Neural Information Processing Systems}, pages
  6713--6723.

\bibitem[{Guu et~al.(2018)Guu, Hashimoto, Oren, and
  Liang}]{guu-etal-2018-generating}
Kelvin Guu, Tatsunori~B. Hashimoto, Yonatan Oren, and Percy Liang. 2018.
\newblock \href {https://doi.org/10.1162/tacl_a_00030} {Generating sentences by
  editing prototypes}.
\newblock \emph{Transactions of the Association for Computational Linguistics},
  6:437--450.

\bibitem[{Higgins et~al.(2016)Higgins, Matthey, Pal, Burgess, Glorot,
  Botvinick, Mohamed, and Lerchner}]{higgins2016beta}
Irina Higgins, Loic Matthey, Arka Pal, Christopher Burgess, Xavier Glorot,
  Matthew Botvinick, Shakir Mohamed, and Alexander Lerchner. 2016.
\newblock beta-{VAE}: Learning basic visual concepts with a constrained
  variational framework.
\newblock In \emph{Proceedings of ICLR}.

\bibitem[{Hochreiter and Schmidhuber(1997)}]{hochreiter1997long}
Sepp Hochreiter and J{\"u}rgen Schmidhuber. 1997.
\newblock Long short-term memory.
\newblock \emph{Neural computation}, 9(8):1735--1780.

\bibitem[{Hu et~al.(2017)Hu, Yang, Liang, Salakhutdinov, and
  Xing}]{hu17control}
Zhiting Hu, Zichao Yang, Xiaodan Liang, Ruslan Salakhutdinov, and Eric~P. Xing.
  2017.
\newblock \href {http://proceedings.mlr.press/v70/hu17e.html} {Toward
  controlled generation of text}.
\newblock In \emph{Proceedings of the 34th International Conference on Machine
  Learning}, volume~70 of \emph{Proceedings of Machine Learning Research},
  pages 1587--1596, International Convention Centre, Sydney, Australia. PMLR.

\bibitem[{Iyyer et~al.(2018)Iyyer, Wieting, Gimpel, and
  Zettlemoyer}]{iyyer2018adversarial}
Mohit Iyyer, John Wieting, Kevin Gimpel, and Luke Zettlemoyer. 2018.
\newblock \href {https://doi.org/10.18653/v1/N18-1170} {Adversarial example
  generation with syntactically controlled paraphrase networks}.
\newblock In \emph{Proceedings of the 2018 Conference of the North American
  Chapter of the Association for Computational Linguistics: Human Language
  Technologies, Volume 1 (Long Papers)}, pages 1875--1885. Association for
  Computational Linguistics.

\bibitem[{John et~al.(2018)John, Mou, Bahuleyan, and
  Vechtomova}]{john2018disentangled}
Vineet John, Lili Mou, Hareesh Bahuleyan, and Olga Vechtomova. 2018.
\newblock Disentangled representation learning for non-parallel text style
  transfer.
\newblock \emph{arXiv preprint arXiv:1808.04339}.

\bibitem[{Ke et~al.(2018)Ke, Guan, Huang, and Zhu}]{ke-etal-2018-generating}
Pei Ke, Jian Guan, Minlie Huang, and Xiaoyan Zhu. 2018.
\newblock \href {https://www.aclweb.org/anthology/P18-1139} {Generating
  informative responses with controlled sentence function}.
\newblock In \emph{Proceedings of the 56th Annual Meeting of the Association
  for Computational Linguistics (Volume 1: Long Papers)}, pages 1499--1508,
  Melbourne, Australia. Association for Computational Linguistics.

\bibitem[{Kingma and Welling(2014)}]{Kingma2014}
Diederik~P. Kingma and Max Welling. 2014.
\newblock {A}uto-{E}ncoding {V}ariational {B}ayes.
\newblock In \emph{Proceedings of ICLR}.

\bibitem[{Kumar et~al.(2017)Kumar, Gupta, Chan, Tucker, Hoffmeister, Dreyer,
  Peshterliev, Gandhe, Filiminov, Rastrow et~al.}]{kumar2017just}
Anjishnu Kumar, Arpit Gupta, Julian Chan, Sam Tucker, Bjorn Hoffmeister, Markus
  Dreyer, Stanislav Peshterliev, Ankur Gandhe, Denis Filiminov, Ariya Rastrow,
  et~al. 2017.
\newblock Just {ASK}: building an architecture for extensible self-service
  spoken language understanding.
\newblock In \emph{1st Workshop on Conversational AI at NIPS 2017 (NIPS-WCAI)}.

\bibitem[{Li et~al.(2019)Li, Qiu, Tang, Chen, Zhao, and
  Yan}]{li2019insufficient}
Juntao Li, Lisong Qiu, Bo~Tang, Dongmin Chen, Dongyan Zhao, and Rui Yan. 2019.
\newblock Insufficient data can also rock! learning to converse using smaller
  data with augmentation.
\newblock In \emph{Thirty-Third AAAI Conference on Artificial Intelligence}.

\bibitem[{Li et~al.(2018)Li, Jiang, Shang, and Li}]{li2018paraphrase}
Zichao Li, Xin Jiang, Lifeng Shang, and Hang Li. 2018.
\newblock \href {http://aclweb.org/anthology/D18-1421} {Paraphrase generation
  with deep reinforcement learning}.
\newblock In \emph{Proceedings of the 2018 Conference on Empirical Methods in
  Natural Language Processing}, pages 3865--3878. Association for Computational
  Linguistics.

\bibitem[{Lin(2004)}]{lin2004rouge}
Chin-Yew Lin. 2004.
\newblock \href {http://aclweb.org/anthology/W04-1013} {{ROUGE}: A package for
  automatic evaluation of summaries}.
\newblock In \emph{Text Summarization Branches Out}.

\bibitem[{Ma et~al.(2018)Ma, Sun, Li, Li, Li, and Ren}]{ma2018query}
Shuming Ma, Xu~Sun, Wei Li, Sujian Li, Wenjie Li, and Xuancheng Ren. 2018.
\newblock \href {https://doi.org/10.18653/v1/N18-1018} {Query and output:
  Generating words by querying distributed word representations for paraphrase
  generation}.
\newblock In \emph{Proceedings of the 2018 Conference of the North American
  Chapter of the Association for Computational Linguistics: Human Language
  Technologies, Volume 1 (Long Papers)}, pages 196--206. Association for
  Computational Linguistics.

\bibitem[{Mallinson et~al.(2017)Mallinson, Sennrich, and
  Lapata}]{mallinson2017paraphrase}
Jonathan Mallinson, Rico Sennrich, and Mirella Lapata. 2017.
\newblock \href {http://aclweb.org/anthology/E17-1083} {Paraphrasing revisited
  with neural machine translation}.
\newblock In \emph{Proceedings of the 15th Conference of the European Chapter
  of the Association for Computational Linguistics: Volume 1, Long Papers},
  pages 881--893. Association for Computational Linguistics.

\bibitem[{Manning et~al.(2014)Manning, Surdeanu, Bauer, Finkel, Bethard, and
  McClosky}]{manning-EtAl:2014:P14-5}
Christopher~D. Manning, Mihai Surdeanu, John Bauer, Jenny Finkel, Steven~J.
  Bethard, and David McClosky. 2014.
\newblock \href {http://www.aclweb.org/anthology/P/P14/P14-5010} {The
  {Stanford} {CoreNLP} natural language processing toolkit}.
\newblock In \emph{Association for Computational Linguistics (ACL) System
  Demonstrations}, pages 55--60.

\bibitem[{Miao et~al.(2016)Miao, Yu, and Blunsom}]{miao2016neural}
Yishu Miao, Lei Yu, and Phil Blunsom. 2016.
\newblock \href {http://proceedings.mlr.press/v48/miao16.html} {Neural
  variational inference for text processing}.
\newblock In \emph{Proceedings of The 33rd International Conference on Machine
  Learning}, volume~48 of \emph{Proceedings of Machine Learning Research},
  pages 1727--1736, New York, New York, USA. PMLR.

\bibitem[{Niu and Bansal(2018)}]{niu2018polite}
Tong Niu and Mohit Bansal. 2018.
\newblock \href {http://aclweb.org/anthology/Q18-1027} {Polite dialogue
  generation without parallel data}.
\newblock \emph{Transactions of the Association for Computational Linguistics},
  6:373--389.

\bibitem[{Pandey et~al.(2018)Pandey, Contractor, Kumar, and
  Joshi}]{pandey-etal-2018-exemplar}
Gaurav Pandey, Danish Contractor, Vineet Kumar, and Sachindra Joshi. 2018.
\newblock \href {https://www.aclweb.org/anthology/P18-1123} {Exemplar
  encoder-decoder for neural conversation generation}.
\newblock In \emph{Proceedings of the 56th Annual Meeting of the Association
  for Computational Linguistics (Volume 1: Long Papers)}, pages 1329--1338,
  Melbourne, Australia. Association for Computational Linguistics.

\bibitem[{Papineni et~al.(2002)Papineni, Roukos, Ward, and Zhu}]{papi2002bleu}
Kishore Papineni, Salim Roukos, Todd Ward, and Wei-Jing Zhu. 2002.
\newblock \href {http://aclweb.org/anthology/P02-1040} {{BLEU}: a method for
  automatic evaluation of machine translation}.
\newblock In \emph{Proceedings of the 40th Annual Meeting of the Association
  for Computational Linguistics}.

\bibitem[{Peng et~al.(2019)Peng, Parikh, Faruqui, Dhingra, and
  Das}]{haopeng-text-19}
Hao Peng, Ankur Parikh, Manaal Faruqui, Bhuwan Dhingra, and Dipanjan Das. 2019.
\newblock \href {https://www.aclweb.org/anthology/N19-1263} {Text generation
  with exemplar-based adaptive decoding}.
\newblock In \emph{Proceedings of the 2019 Conference of the North {A}merican
  Chapter of the Association for Computational Linguistics: Human Language
  Technologies, Volume 1 (Long and Short Papers)}, pages 2555--2565,
  Minneapolis, Minnesota. Association for Computational Linguistics.

\bibitem[{Pennington et~al.(2014)Pennington, Socher, and
  Manning}]{pennington2014glove}
Jeffrey Pennington, Richard Socher, and Christopher~D. Manning. 2014.
\newblock \href {http://www.aclweb.org/anthology/D14-1162} {Glove: Global
  vectors for word representation}.
\newblock In \emph{Empirical Methods in Natural Language Processing (EMNLP)},
  pages 1532--1543.

\bibitem[{Plank et~al.(2016)Plank, S{\o}gaard, and
  Goldberg}]{plank2016multilingual}
Barbara Plank, Anders S{\o}gaard, and Yoav Goldberg. 2016.
\newblock \href {https://doi.org/10.18653/v1/P16-2067} {Multilingual
  part-of-speech tagging with bidirectional long short-term memory models and
  auxiliary loss}.
\newblock In \emph{Proceedings of the 54th Annual Meeting of the Association
  for Computational Linguistics (Volume 2: Short Papers)}, pages 412--418.
  Association for Computational Linguistics.

\bibitem[{Prakash et~al.(2016)Prakash, Hasan, Lee, Datla, Qadir, Liu, and
  Farri}]{prakash2016neural}
Aaditya Prakash, Sadid~A. Hasan, Kathy Lee, Vivek Datla, Ashequl Qadir, Joey
  Liu, and Oladimeji Farri. 2016.
\newblock \href {http://aclweb.org/anthology/C16-1275} {Neural paraphrase
  generation with stacked residual {LSTM} networks}.
\newblock In \emph{Proceedings of COLING 2016, the 26th International
  Conference on Computational Linguistics: Technical Papers}, pages 2923--2934.
  The COLING 2016 Organizing Committee.

\bibitem[{Quirk et~al.(2004)Quirk, Brockett, and Dolan}]{quirk2004mono}
Chris Quirk, Chris Brockett, and William Dolan. 2004.
\newblock \href {http://aclweb.org/anthology/W04-3219} {Monolingual machine
  translation for paraphrase generation}.
\newblock In \emph{Proceedings of the 2004 Conference on Empirical Methods in
  Natural Language Processing}.

\bibitem[{Rei(2017)}]{rei2017semi}
Marek Rei. 2017.
\newblock \href {https://doi.org/10.18653/v1/P17-1194} {Semi-supervised
  multitask learning for sequence labeling}.
\newblock In \emph{Proceedings of the 55th Annual Meeting of the Association
  for Computational Linguistics (Volume 1: Long Papers)}, pages 2121--2130.
  Association for Computational Linguistics.

\bibitem[{Semeniuta et~al.(2017)Semeniuta, Severyn, and Barth}]{seme2017hybrid}
Stanislau Semeniuta, Aliaksei Severyn, and Erhardt Barth. 2017.
\newblock \href {https://doi.org/10.18653/v1/D17-1066} {A hybrid convolutional
  variational autoencoder for text generation}.
\newblock In \emph{Proceedings of the 2017 Conference on Empirical Methods in
  Natural Language Processing}, pages 627--637. Association for Computational
  Linguistics.

\bibitem[{Serban et~al.(2017)Serban, Ororbia, Pineau, and
  Courville}]{serban2017piecewise}
Iulian~Vlad Serban, Alexander~G. Ororbia, Joelle Pineau, and Aaron Courville.
  2017.
\newblock \href {https://doi.org/10.18653/v1/D17-1043} {Piecewise latent
  variables for neural variational text processing}.
\newblock In \emph{Proceedings of the 2017 Conference on Empirical Methods in
  Natural Language Processing}, pages 422--432. Association for Computational
  Linguistics.

\bibitem[{Shen et~al.(2019)Shen, Celikyilmaz, Zhang, Chen, Wang, Gao, and
  Carin}]{shen2019generating}
Dinghan Shen, Asli Celikyilmaz, Yizhe Zhang, Liqun Chen, Xin Wang, Jianfeng
  Gao, and Lawrence Carin. 2019.
\newblock \href {http://arxiv.org/abs/1902.00154} {Towards generating long and
  coherent text with multi-level latent variable models}.

\bibitem[{Shen et~al.(2017)Shen, Lei, Barzilay, and Jaakkola}]{shen2017style}
Tianxiao Shen, Tao Lei, Regina Barzilay, and Tommi Jaakkola. 2017.
\newblock Style transfer from non-parallel text by cross-alignment.
\newblock In \emph{Advances in Neural Information Processing Systems}, pages
  6830--6841.

\bibitem[{Toutanova et~al.(2003)Toutanova, Klein, Manning, and
  Singer}]{krist20013feature}
Kristina Toutanova, Dan Klein, Christopher~D. Manning, and Yoram Singer. 2003.
\newblock \href {http://aclweb.org/anthology/N03-1033} {Feature-rich
  part-of-speech tagging with a cyclic dependency network}.
\newblock In \emph{Proceedings of the 2003 Human Language Technology Conference
  of the North American Chapter of the Association for Computational
  Linguistics}.

\bibitem[{Wang et~al.(2019)Wang, Hu, Yang, Shi, Xu, and Xing}]{wang2019toward}
Wentao Wang, Zhiting Hu, Zichao Yang, Haoran Shi, Frank Xu, and Eric Xing.
  2019.
\newblock Toward unsupervised text content manipulation.
\newblock \emph{arXiv preprint arXiv:1901.09501}.

\bibitem[{Weston et~al.(2018)Weston, Dinan, and
  Miller}]{weston-etal-2018-retrieve}
Jason Weston, Emily Dinan, and Alexander Miller. 2018.
\newblock \href {https://www.aclweb.org/anthology/W18-5713} {Retrieve and
  refine: Improved sequence generation models for dialogue}.
\newblock In \emph{Proceedings of the 2018 {EMNLP} Workshop {SCAI}: The 2nd
  International Workshop on Search-Oriented Conversational {AI}}, pages 87--92,
  Brussels, Belgium. Association for Computational Linguistics.

\bibitem[{Wieting et~al.(2016)Wieting, Bansal, Gimpel, and
  Livescu}]{wieting-16-full}
John Wieting, Mohit Bansal, Kevin Gimpel, and Karen Livescu. 2016.
\newblock Towards universal paraphrastic sentence embeddings.
\newblock In \emph{Proceedings of ICLR}.

\bibitem[{Wieting and Gimpel(2018)}]{para-nmt-acl-18}
John Wieting and Kevin Gimpel. 2018.
\newblock \href {http://aclweb.org/anthology/P18-1042} {{ParaNMT-50M}: Pushing
  the limits of paraphrastic sentence embeddings with millions of machine
  translations}.
\newblock In \emph{Proceedings of the 56th Annual Meeting of the Association
  for Computational Linguistics (Volume 1: Long Papers)}, pages 451--462.
  Association for Computational Linguistics.

\bibitem[{Wiseman et~al.(2017)Wiseman, Shieber, and
  Rush}]{wiseman2017challenges}
Sam Wiseman, Stuart Shieber, and Alexander Rush. 2017.
\newblock Challenges in data-to-document generation.
\newblock In \emph{Proceedings of the 2017 Conference on Empirical Methods in
  Natural Language Processing}, pages 2253--2263.

\bibitem[{Wiseman et~al.(2018)Wiseman, Shieber, and
  Rush}]{wiseman2018templates}
Sam Wiseman, Stuart Shieber, and Alexander Rush. 2018.
\newblock \href {http://aclweb.org/anthology/D18-1356} {Learning neural
  templates for text generation}.
\newblock In \emph{Proceedings of the 2018 Conference on Empirical Methods in
  Natural Language Processing}, pages 3174--3187. Association for Computational
  Linguistics.

\bibitem[{Xu and Durrett(2018)}]{xu2018spherical}
Jiacheng Xu and Greg Durrett. 2018.
\newblock \href {http://aclweb.org/anthology/D18-1480} {Spherical latent spaces
  for stable variational autoencoders}.
\newblock In \emph{Proceedings of the 2018 Conference on Empirical Methods in
  Natural Language Processing}, pages 4503--4513. Association for Computational
  Linguistics.

\bibitem[{Zhang and Shasha(1989)}]{zhang1989simple}
Kaizhong Zhang and Dennis Shasha. 1989.
\newblock Simple fast algorithms for the editing distance between trees and
  related problems.
\newblock \emph{SIAM journal on computing}, 18(6):1245--1262.

\bibitem[{Zhao et~al.(2018)Zhao, Kim, Zhang, Rush, and
  LeCun}]{zhao2018adversarially}
Junbo Zhao, Yoon Kim, Kelly Zhang, Alexander~M Rush, and Yann LeCun. 2018.
\newblock {A}dversarially {R}egularized {A}utoencoders.
\newblock In \emph{Proceedings of ICML}.

\bibitem[{Zhao et~al.(2017)Zhao, Zhao, and Eskenazi}]{zhao2017learning}
Tiancheng Zhao, Ran Zhao, and Maxine Eskenazi. 2017.
\newblock Learning discourse-level diversity for neural dialog models using
  conditional variational autoencoders.
\newblock In \emph{Proceedings of the 55th Annual Meeting of the Association
  for Computational Linguistics (Volume 1: Long Papers)}, volume~1, pages
  654--664.

\bibitem[{Zhou and Neubig(2017)}]{zhou2017multi}
Chunting Zhou and Graham Neubig. 2017.
\newblock \href {https://doi.org/10.18653/v1/P17-1029} {Multi-space variational
  encoder-decoders for semi-supervised labeled sequence transduction}.
\newblock In \emph{Proceedings of the 55th Annual Meeting of the Association
  for Computational Linguistics (Volume 1: Long Papers)}, pages 310--320.
  Association for Computational Linguistics.

\end{thebibliography}
\bibliographystyle{acl_natbib}

\appendix

\section{Appendices}

\subsection{Hyperparameters}

We use 100 dimensional word embeddings in both encoders and 100 dimensional word embeddings for the decoder. These word embeddings are all initialized by GloVe vectors~\cite{pennington2014glove}. The syntactic encoder uses 100 dimensions per direction and the decoder is a 100 dimensional unidirectional LSTM. When performing early stopping, we use greedy decoding. During testing, we use beam search with size of 10.

\end{document}